\documentclass[final]{cvpr}

\usepackage{times}
\usepackage{epsfig}
\usepackage{graphicx}
\usepackage{amsmath}
\usepackage{amssymb}
\usepackage{mathtools}

\usepackage[pagebackref=true,breaklinks=true,colorlinks,bookmarks=false]{hyperref}

\def\cvprPaperID{8778} 
\def\confYear{CVPR 2021}
\begin{document}

\title{DiNTS: Differentiable Neural Network Topology Search\\for 3D Medical Image Segmentation}


\author{Yufan He\textsuperscript{1} \quad Dong Yang\textsuperscript{2}  \quad Holger Roth\textsuperscript{2}  \quad Can Zhao\textsuperscript{2}  \quad Daguang Xu\textsuperscript{2}\\
\textsuperscript{1}Johns Hopkins University \quad \textsuperscript{2}NVIDIA}

\maketitle

\begin{abstract}
   Recently, neural architecture search~(NAS) has been applied to automatically search high-performance networks for medical image segmentation.  The NAS search space usually contains a network topology level~(controlling connections among cells with different spatial scales) and a cell level~(operations within each cell). Existing methods either require long searching time for large-scale 3D image datasets, or are limited to pre-defined topologies (such as U-shaped or single-path) . In this work, we focus on three important aspects of NAS in 3D medical image segmentation: flexible multi-path network topology, high search efficiency, and budgeted GPU memory usage. A novel differentiable search framework is proposed to support fast gradient-based search within a highly flexible network topology search space. The discretization of the searched optimal continuous model in differentiable scheme may produce a sub-optimal final discrete model~(discretization gap). Therefore, we propose a topology loss to alleviate this problem. In addition, the GPU memory usage for the searched 3D model is limited with budget constraints during search. Our \textbf{Di}fferentiable \textbf{N}etwork \textbf{T}opology \textbf{S}earch scheme~(\textbf{DiNTS}) is evaluated on the Medical Segmentation Decathlon (MSD) challenge, which contains ten challenging segmentation tasks. Our method achieves the state-of-the-art performance and the top ranking on the MSD challenge leaderboard.
\end{abstract}

\section{Introduction}
\label{s:intro}
\begin{figure}[t]
\begin{center}
\includegraphics[width=1\linewidth]{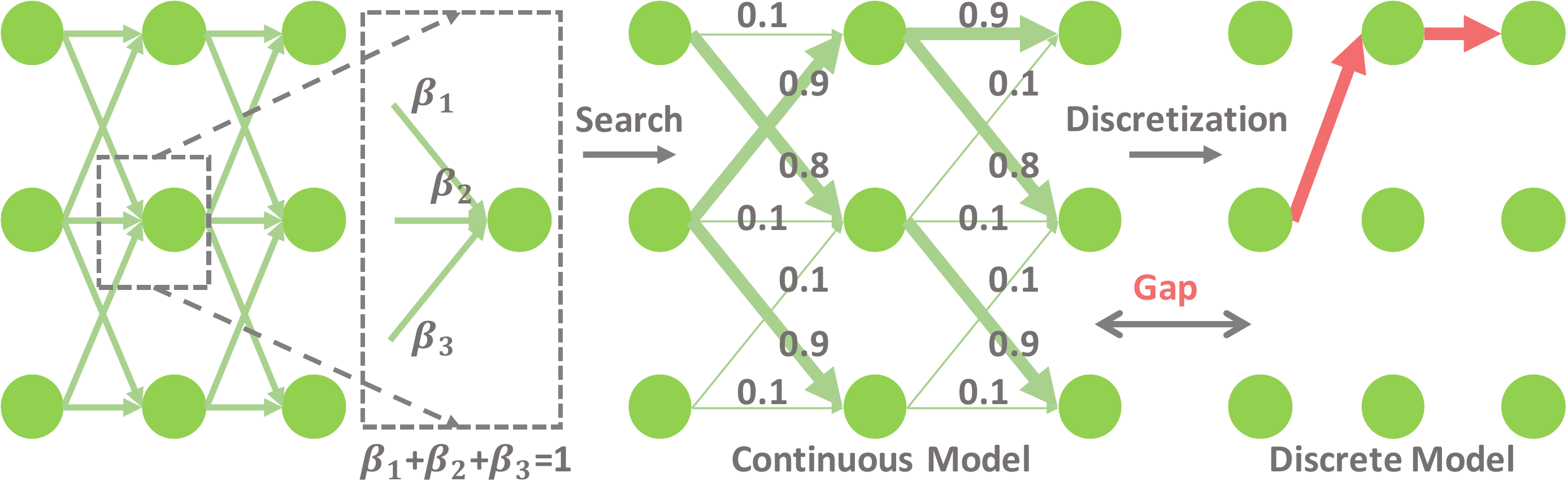}
\end{center}
  \caption{Limitations of existing differentiable topology search formulation. E.g. in Auto-DeepLab~\cite{liu2019auto}, each edge in the topology search space is given a probability $\beta$. The probabilities of input edges to a node sum to one, which means only one input edge for each node would be selected. A single-path discrete model~(red path) is extracted from the continuous searched model. This can result in a large ``discretization gap'' between the feature flow of the searched continuous model and the final discrete model.}
\label{fig:ad}
\end{figure}
Automated medical image segmentation is essential for many clinical applications like finding new biomarkers and monitoring disease progression. The recent developments in deep neural network architectures have achieved great performance improvements in image segmentation. Manually designed networks, like U-Net~\cite{ronneberger2015unet}, have been widely used in different tasks. However, the diversity of medical image segmentation tasks could be extremely high since the image characteristics \& appearances can be completely distinct for different modalities and the presentation of diseases can vary considerably. This makes the direct application of even a successful network like U-Net~\cite{ronneberger2015unet} to a new task less likely to be optimal. 

The neural architecture search~(NAS) algorithms~\cite{zoph2016nas} have been proposed to automatically discover the optimal architectures within a search space. The NAS search space for segmentation usually contains two levels: network topology level and cell level. The network topology controls the connections among cells and decides the flow of the feature maps across different spatial scales. The cell level decides the specific operations on the feature maps. A more flexible search space has more potential to contain better performing architectures. 

In terms of the search methods in finding the optimal architecture from the search space,  evolutionary or reinforcement learning-based~\cite{zoph2016nas,real2019regularized} algorithms are usually time consuming. C2FNAS~\cite{yu2020c2fnas} takes 333 GPU days to search one 3D segmentation network using the evolutionary-based methods, which is too computationally expensive for common use cases. Differentiable architecture search~\cite{liu2018darts} is much more efficient and Auto-DeepLab~\cite{liu2019auto} is the first work to apply differentiable search for segmentation network topology. However, Auto-DeepLab's differentiable formulation limits the searched network topology. As shown in Fig.~\ref{fig:ad}, this formulation assumes that only one input edge would be kept for each node.  Its final searched model only has a single path from input to output which limits its complexity. Our first goal is to propose a new differentiable scheme to support more complex topologies in order to find novel architectures with better performance. 

Meanwhile, the differentiable architecture search suffers from the ``discretization gap'' problem~\cite{chen2019progressive,tian2020discretization}. The discretization of the searched optimal continuous model may produce a sub-optimal discrete final architecture and cause a large performance gap. As shown in Fig.~\ref{fig:ad}, the gap comes from two sides: 1) the searched continuous model is not binary, thus some operations/edges with small but non-zero probabilities are discarded during the discretization step; 2) the discretization algorithm has topology constraints~(e.g. single-path), thus edges causing infeasible topology are not allowed even if they have large probabilities in the continuous model. Alleviating the first problem by encouraging a binarized model during search has been explored~\cite{chu2019fair,tian2020discretization,nayman2019xnas}. However, alleviating the second problem requires the search to be aware of the discretization algorithm and topology constraints. In this paper, we propose a topology loss in search stage and a topology guaranteed discretization algorithm to mitigate this problem.  

In medical image analysis, especially for some longitudinal analysis tasks, high input image resolution and large patch size are usually desired to capture miniscule longitudinal changes. Thus, large GPU memory usage is a major challenge for training with large high resolution 3D images. Most NAS algorithms with computational constraints focus on latency~\cite{cai2018proxylessnas,chen2019fasterseg,li2019partial,Shaw_2019_ICCV} for real-time applications. However, real-time inference often is not a major concern compared to the problem caused by huge GPU memory usage in 3D medical image analysis. In this paper, we propose additional GPU memory constraints in the search stage to limit the GPU usage needed for retraining the searched model.

We validate our method on the Medical Segmentation Decathlon~(MSD) dataset~\cite{simpson2019large}  which contains 10 representative 3D medical segmentation tasks covering different anatomies and imaging modalities. We achieve state-of-the-art results while only takes 5.8 GPU days~(recent C2FNAS~\cite{yu2020c2fnas} takes 333 GPU days on the same dataset). Our contributions can be summarized as: 
\begin{itemize}
  \item We propose a novel \textbf{Di}fferentiable \textbf{N}etwork \textbf{T}opology \textbf{S}earch scheme \textbf{DiNTS}, which supports more flexible topologies and joint two-level search.
  \item We propose a topology guaranteed discretization algorithm and a discretization aware topology loss for the search stage to minimize the discretization gap.
  \item We develop a memory usage aware search method which is able to search 3D networks with different GPU memory requirements.
  \item We achieve the new state-of-the-art results and top ranking in the MSD challenge leaderboard while only taking 1.7\% of the search time compared to the NAS-based C2FNAS~\cite{yu2020c2fnas}.
\end{itemize}


\begin{figure*}[t]
\begin{center}
\includegraphics[width=1\linewidth]{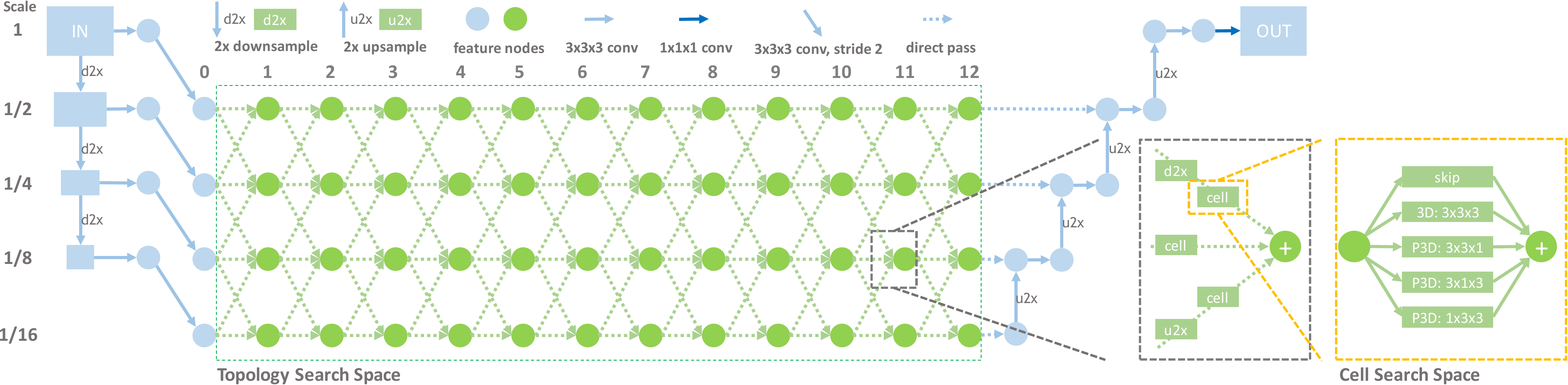}
\end{center}
  \caption{Our search space contains $L$=12 layers. The blue edges are the stem containing pre-defined operations. The cell operations are defined on the edges while the nodes are feature maps. Edges in the topology search space that are selected for features to flow from input to output form a candidate network topology. Each edge in the search space includes a cell which contains $O$=5 operations to select from. A downsample/upsample edge also contains a 2$\times$ downsample/upsample operation.}
\label{fig:ss}
\end{figure*}
\section{Related Work}

\subsection{Medical Image Segmentation}
 Medical image segmentation faces some unique challenges like lacking manual labels and vast memory usage for processing 3D high resolution images. Compared to networks used in natural images like DeepLab~\cite{chen2018encoder} and PSPNet~\cite{zhao2017psp}, 2D/3D UNet~\cite{ronneberger2015unet,cciccek20163d} is better at preserving fine details and memory friendly when applied to 3D images.
 VNet~\cite{milletari2016v} improves 3D UNet with residual blocks. UNet++~\cite{zhou2019unet++} uses dense blocks~\cite{huang2017densely} to redesign skip connections. H-DenseUNet~\cite{li2018h} combines 2D and 3D UNet to save memory. nnUNet~\cite{isensee2019nnunet} ensembles 2D, 3D, and cascaded 3D UNet and achieves state-of-the-art results on a variety of medical image segmentation benchmarks. 
\subsection{Neural Architecture Search}
Neural architecture search~(NAS) focuses on designing network automatically. The work in NAS can be categorized into three dimensions: search space, search method and performance estimation~\cite{elsken2018neural}. The search space defines what architecture can be searched, which can be further divided into network topology level and cell level. For image classification, \cite{liu2018darts,zoph2018learning,liu2018progressive,real2019regularized,pham2018efficient,gu2020dots} focus on searching optimal cells and apply a pre-defined network topology while \cite{fang2020densely,xie2019exploring} perform search on the network topology. In segmentation, Auto-DeepLab~\cite{liu2019auto} uses a highly flexible search space while FasterSeg~\cite{chen2019fasterseg} proposes a low latency two level search space. Both perform a joint two-level search.  In medical image segmentation, NAS-UNet~\cite{weng2019nasunet}, V-NAS~\cite{zhu2019vnas} and Kim et al~\cite{kim2019scalable} search cells and apply it to a U-Net-like topology. C2FNAS~\cite{yu2020c2fnas} searches 3D network topology in a U-shaped space and then searches the operation for each cell. MS-NAS~\cite{yan2020ms} applies PC-Darts~\cite{xu2019pc} and Auto-DeepLab's formulation to 2D medical images.

Search method and performance estimation focus on finding the optimal architecture from the search space. Evolutionary and reinforcement learning has been used in~\cite{zoph2016nas,real2019regularized} but those methods require extremely long search time. Differentiable methods~\cite{liu2018darts,liu2019auto} relax the discrete architecture into continuous representations and allow direct gradient based search. This is magnitudes faster and has been applied in various NAS works~\cite{liu2018darts,liu2019auto,xu2019pc,zhu2019vnas,yan2020ms}. However, converting the continuous representation back to the discrete architecture causes the ``discretization gap''. To solve this problem, FairDARTS~\cite{chu2019fair} and Tian et al~\cite{tian2020discretization} proposed zero-one loss and entropy loss respectively to push the continuous representation close to binary. Some works~\cite{nayman2019xnas, hu2020dsnas} use temperature annealing to achieve the same goal. Another problem of the differentiable method is the large memory usage during search stage. PC-DARTS~\cite{xu2019pc} uses partial channel connections to reduce memory, while Auto-DeepLab~\cite{liu2019auto} reduces the filter number at search stage. It's a common practice to retrain the searched model while increasing the filter number, batch size, or patch size to gain better performance. But for 3D medical image segmentation, the change of retraining scheme~(e.g. transferring to a new task which requires larger input size) can still cause out-of-memory problem. Most NAS work has been focused on searching architecture with latency constraints~\cite{cai2018proxylessnas,chen2019fasterseg,li2019partial,Shaw_2019_ICCV}, while only a few considered memory as a constraint. Mem-NAS~\cite{liu2020memnas} uses a growing and trimming framework to constrain the inference GPU memory but does not allow integration in a differentiable scheme.   

\section{Method}

\begin{figure}
\begin{center}
\includegraphics[width=1\linewidth]{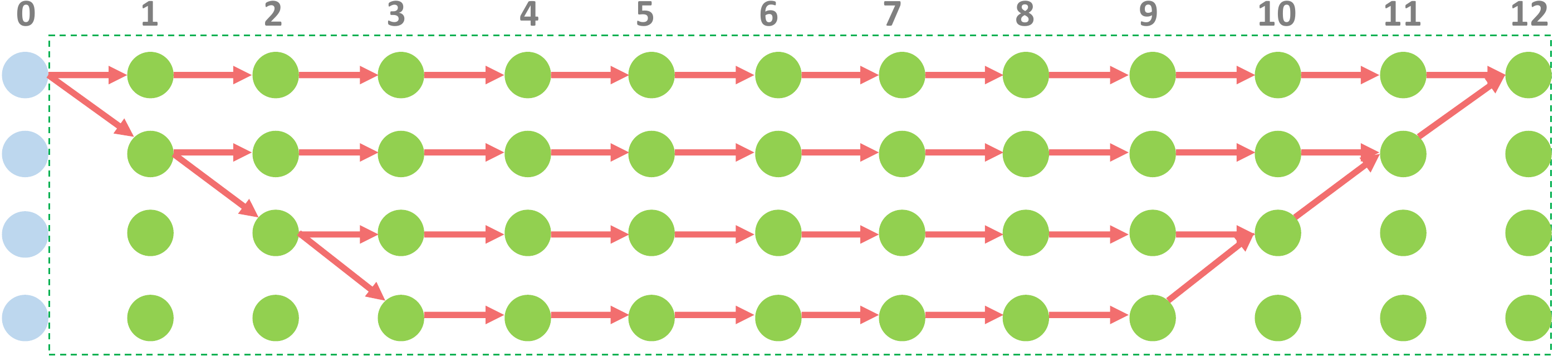}
(a) Multi-path topology: UNet~\cite{ronneberger2015u}
\includegraphics[width=1\linewidth]{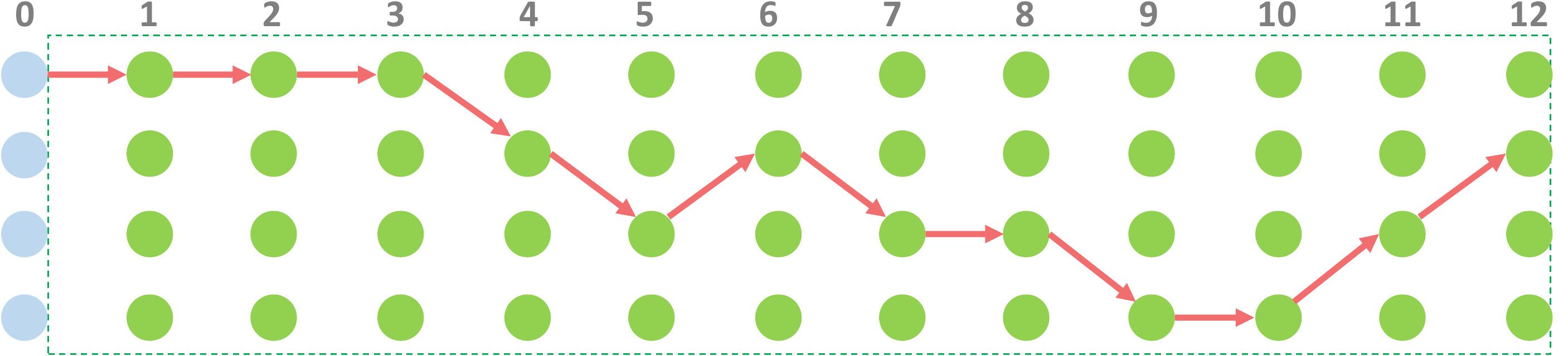}
(b) Single-path topology: Auto-DeepLab~\cite{liu2019auto}
\includegraphics[width=1\linewidth]{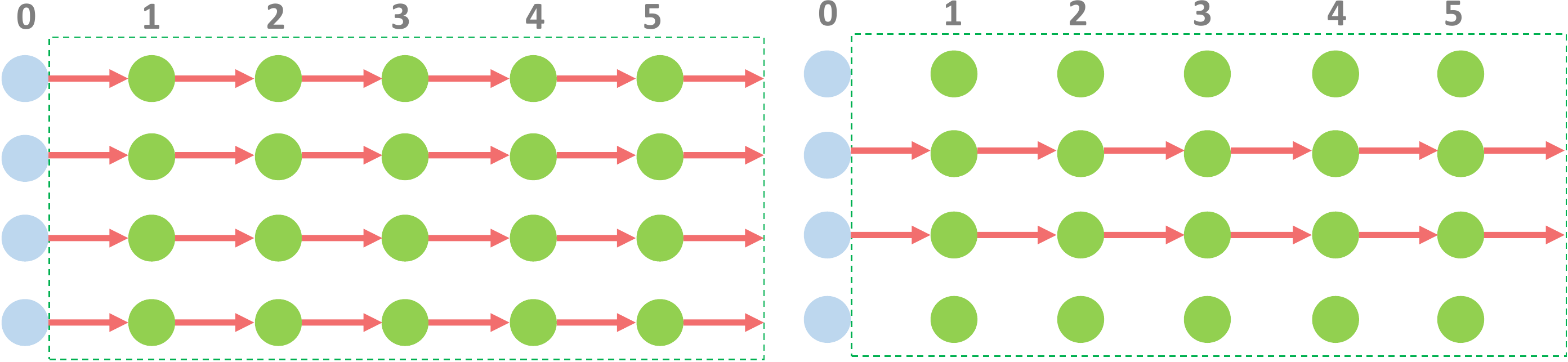}
(c) Multi-resolution input~\cite{Lin_2017_CVPR} and Input selection
\end{center}
  \caption{Our search space covers a variety of topologies (single-path, multi-path) and can select input resolutions.}
\label{fig:ss_example}
\end{figure}

\subsection{Network Topology Search Space}
Inspired by Auto-Deeplab~\cite{liu2019auto} and \cite{li2020learning}, we propose a search space with fully connected edges between adjacent resolutions~(2$\times$ higher, 2$\times$ lower or the same) from adjacent layers as shown in Fig.~\ref{fig:ss}. A stack of multi-resolution images are generated by down-sampling the input image by 1/2, 1/4, 1/8 along each axis. Together with the original image, we use four $3\times 3 \times 3$ 3D convolutions with stride 2 to generate multi-resolution features (layer 0 in Fig.~\ref{fig:ss}) to the following search space. The search space has $L$ layers and each layer consists of feature nodes~(green nodes) from $D$=4 resolutions and $E$=3$D$-2 candidate input edges~(dashed green edges). Each edge contains a cell operation, and a upsample/downsample operation (factor 2) is used before the cell if the edge is an upsample/downsample edge. A feature node is the summation of the output features from each input edge. Compared to Auto-DeepLab~\cite{liu2019auto}, our search space supports searching for input image scales and complex multi-path topologies, as shown in Fig.~\ref{fig:ss_example}. As for multi-path topology, MS-NAS~\cite{yan2020ms} discretizes and combines multiple single-path models searched from Auto-DeepLab's framework, but the search is still unaware of the discretization thus causing the gap. \cite{li2020learning} also supports multi-path topology, but \cite{li2020learning} is more about feature routing in a ``fully connected'' network, not a NAS method.

\subsection{Cell Level Search Space}
\label{s:cell}
We define a cell search space to be a set of basic operations where the input and output feature maps have the same spatial resolution. The cell search space in DARTS~\cite{liu2018darts} and Auto-Deeplab~\cite{liu2019auto} contains multiple blocks and the connections among those blocks can also be searched. However, the searched cells are repeated over all the cells in the network topology level. Similar to C2FNAS~\cite{yu2020c2fnas}, our algorithm searches the operation of each cell independently, with one operation selected from the following:
\begin{tabular}{l}
    (1) skip connection \quad\quad\quad\, (2) 3x3x3 3D convolution \\
    (3) P3D 3x3x1: 3x3x1 followed by 1x1x3 convolution \\
    (4) P3D 3x1x3: 3x1x3 followed by 1x3x1 convolution \\
    (5) P3D 1x3x3: 1x3x3 followed by 3x1x1 convolution
\end{tabular}
 P3D represents pseudo 3D~\cite{qiu2017p3d} and has been used in V-NAS~\cite{zhu2019vnas}. A cell also includes ReLU activation and Instance Normalization~\cite{ulyanov2016instance} which are used before and after those operations respectively~(except for skip connection). The cell operations do not include multi-scale feature aggregation operations like atrous convolution and pooling. The feature spatial changes are performed by the upsample/downsample operations in the edges searched from the topology level.

\subsection{Continuous Relaxation and Discretization}
\subsubsection{Preliminaries}
\label{s:pre}
 We briefly recap the relaxation in DARTS~\cite{liu2018darts}. NAS tries to select one from $N$ candidate operations $O_1, O_2, \cdots, O_N$ for each computational node. Each operation $O_i$ is paired with a trainable parameter $\alpha_i$ where
 $\sum_{i=1}^N\alpha_i=1$, $\alpha_i\ge 0$, and the output feature $x_{out}=\sum_{i=1}^N\alpha_iO_i(x_{in})$, where $x_{in}$ is the input feature. Thus, the discrete operation is relaxed by the continuous representation $\alpha$ which can be optimized using gradient descent. After optimization,  $O_i$ with larger $\alpha_i$ is more important and will be selected. However, a small $\alpha_j$~(as long as $\alpha_j \neq 0$) can still make a significant difference on $x_{out}$ and following layers. Therefore, directly discarding non-zero operations will lead to the discretization gap.

Auto-DeepLab~\cite{liu2019auto} extends this idea to edge selection in network topology level. As illustrated in Fig.~\ref{fig:ad}, every edge is paired with a trainable parameter $\beta$~($0\ge \beta \ge 1$), and parameters paired with edges that pointed to the same feature node sum to one. This is based on an assumption that ``one input edge for each node'' because the input edges to a node are competing with each other. After discretization, a single path is kept while other edges, even with a large $\beta$, are discarded. This means the feature flow in the searched continuous model has a significant gap with the feature flow in the final discrete model. The single-path topology limitation comes from the previous assumption for topology level relaxation while the gap comes from the unawareness of the discretization in the search stage, such that edges with large probabilities can be discarded due to topology. 
\begin{figure}[]
\begin{center}
\includegraphics[width=1\linewidth]{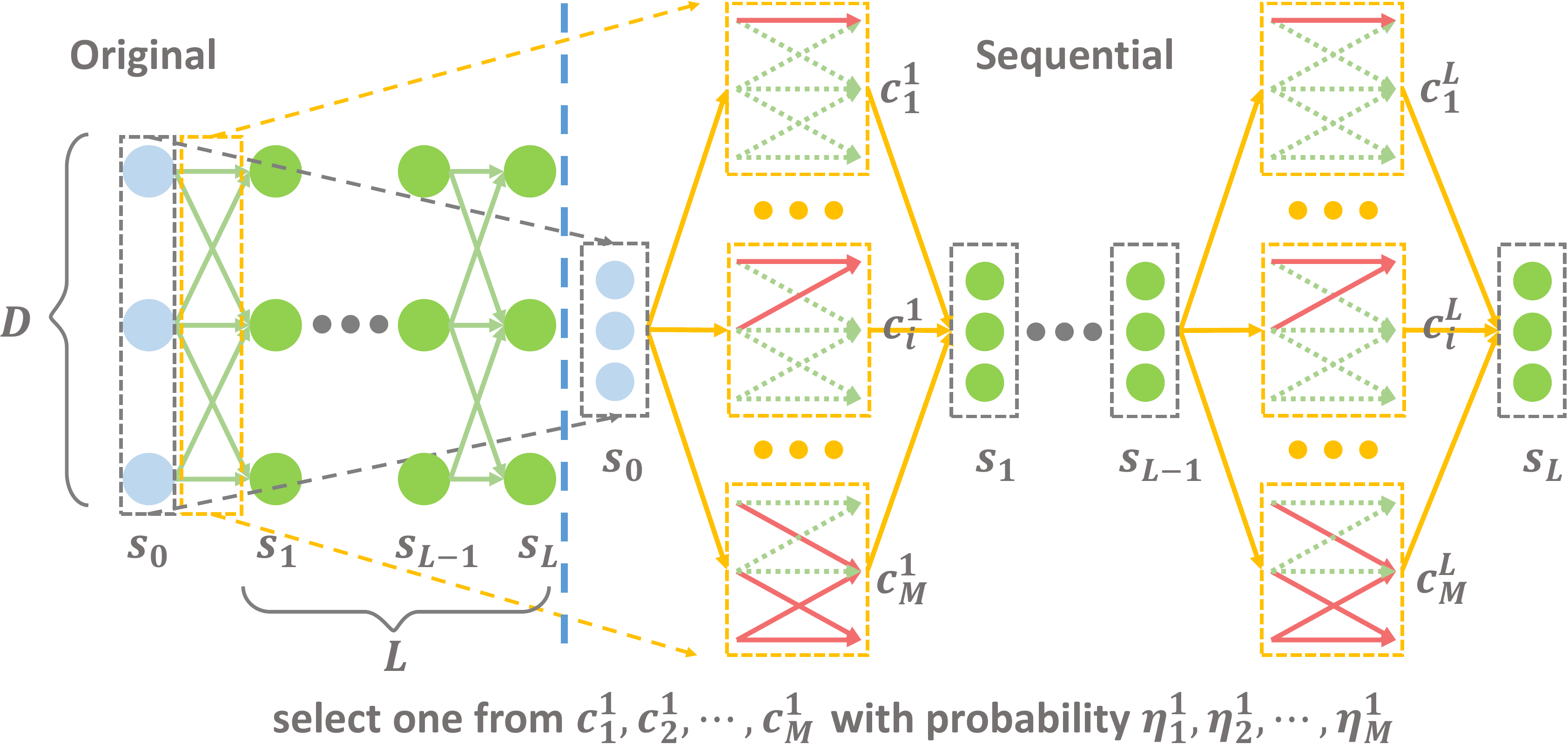}
\end{center}
 \caption{The feature nodes at the same layer $i$ are combined as a super node $s_i$. A set of selected edges~(e.g. red edges in a dashed yellow block) that connects $s_{i-1}$ and $s_i$ is a ``connection''. For $E$ edges, there are $M=2^E-1$ connection patterns. Topology search becomes selecting one connection pattern to connect adjacent super nodes sequentially.}
\label{fig:sn}
\end{figure}

\subsubsection{Sequential Model with Super Feature Node}
We propose a network topology relaxation framework which converts the multi-scale search space into a sequential space using ``Super Feature Node''.
For a search space with $L$ layers and $D$ resolution levels, these $D$ feature nodes in the same layer $i$ are combined as a super feature node $s_i$ and features flow sequentially from these $L$ super nodes as shown in Fig.~\ref{fig:sn}. There are $E$=3$D$-2 candidate input edges to each super node and the topology search is to select an optimal set of input edges for each super node. We define a connection pattern as a set of selected edges and there are $M=2^E-1$ feasible candidate connection patterns. The $j$-th connection pattern $cp_j$ is an indication vector of length $E$, where $cp_j(e)=1$, if $e$-th edge is selected in $j$-th pattern.

We define the input connection operation to $s_i$ with connection pattern $cp_j$ as $c^i_j$. $cp_j$ defines $c^i_j$'s topology while $c^i_j$ also includes cell operations on the selected edges in $cp_j$. $c^i_j,c^{i+1}_k$ means the input/output connection patterns for $s_i$ are $cp_j,cp_k$ respectively. 
Under this formulation, the topology search becomes selecting an input connection pattern for each super node and the competition is among all $M$ connection patterns, not among edges. We associate a variable $\eta^i_j$ to the connection operation $c^i_j$ for every $s_i$ and every pattern $j$. Denote the input features at layer 0 as $s_0$, we have a sequential feature flow equation:
\begin{gather}
\label{eq:flow}
    s_{i} = \sum_{j=1}^M(\eta^i_j*c^i_j(s_{i-1})) \quad i=1\cdots,L \\ \nonumber
    \sum_{j=1}^M\eta^i_j=1, \eta_j\ge 0 \quad \forall i,j 
\end{gather}
\begin{gather}
\label{eq:eta}
    \eta^i_j =  \frac{\prod_{e=1}^E (1-p^i_e)^{1-cp_j(e)}(p^i_e)^{cp_j(e)}}{\sum_{j=1}^M \prod_{e=1}^E (1-p^i_e)^{1-cp_j(e)}(p^i_e)^{cp_j(e)}} \\ \nonumber
    0 \le p^i_e \le 1 \quad \forall i,e 
\end{gather}
However, $M$ is growing exponentially with $D$. To reduce the architecture parameters, we parameterize $\eta^i_j$ with a set of edge probability parameters $p^i_e$, $e$=$1,\cdots,E$ in Eq.~\ref{eq:eta}. For a search space with $L$=12 layers and $D$=4, the network topology parameter number is reduced from $M\times L=1023\times12$ to $E\times L=10\times12$. Under this formulation, the probability $\eta$ of connections are highly correlated. If an input edge $e$ to $s_i$ has low probability, all the candidate patterns to $s_i$ with $e$ selected will have lower probabilities.

For cell operation relaxation, we use the method in Sec.~\ref{s:pre}. Each cell on the input edge $e$ to $s_i$ has its own cell architecture parameters $\alpha^{i,e}_1, \alpha^{i,e}_2, \cdots, \alpha^{i,e}_N$ and will be optimized. Notice the $c^i_j$ in Eq.~\ref{eq:flow}. contains the cell operations defined on the selected edges, and it contains relaxed cell architecture parameters $\alpha$. Thus we can perform gradient based search for topology and cell levels jointly.

\subsubsection{Discretization with Topology Constraints}
\label{s:dis}
\begin{figure}
\begin{center}
\includegraphics[width=1\linewidth]{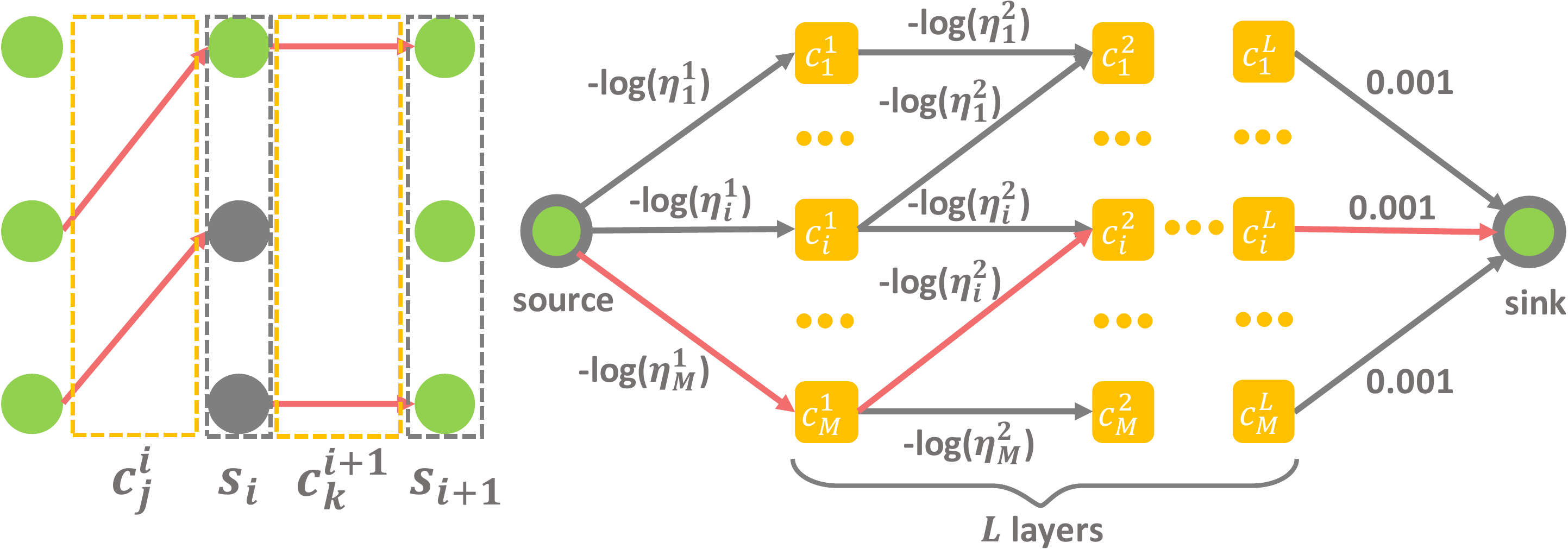}
\end{center}
\caption{\emph{Left:} The gray feature nodes are topologically infeasible, thus connection pattern index $k$ is not in $j$'s feasible set, $k\notin \mathcal{F}(j)$.
\emph{Right:} A directed graph $\mathcal{G}$ which contains $L\times M$+2 nodes. A node $c^i_j$~(yellow block) is connected with $c^{i+1}_k$ and $c^{i-1}_m$ if $j\in \mathcal{F}(m)$ and $k\in \mathcal{F}(j)$. The cost of edges directed to $c^i_j$ is $\text{-log}\eta^i_j$. The source connects to all first layer nodes and all $L$-th layer nodes connect to the sink~(edge cost is a constant value).  Those $L$ nodes on the shortest path from source to sink~(red path) in  $\mathcal{G}$ represent the optimal feasible connection operations~(final architecture).}
\label{fig:tp}
\end{figure}

After training, the final discrete architecture is derived from the optimized continuous architecture representation $\eta$~(derived from $p^i_e$) and $\alpha$. $\eta^j_i$ represents the probability of using input connection pattern $cp^i_j$ for super node $s_i$. Since the network topology search space is converted into a sequential space, the easiest way for topology discretization is to select $cp_j$ with the maximum $\eta^i_j$. However, the topology may not be feasible. We define topology infeasibility as:
\begin{center}
\emph{``a feature node has an input edge but no output edge or has an output edge but no input edge''}. 
\end{center}
The gray feature nodes in Fig.~\ref{fig:tp} indicate infeasible topology. Therefore, we cannot select $cp_j$ and $cp_k$ as $s_i$'s input/output connection patterns even if they have the largest probabilities. For every connection pattern $cp_j$, we generate a feasible set $\mathcal{F}(j)$. If a super node with input pattern $j$ and output pattern $k$ is feasible~(all feature nodes of the super node are topologically feasible), then $k\in \mathcal{F}(j)$. Denote the array of selected input connection pattern indexes for these $L$ super nodes as $I$, and the topology discretization can be performed by sampling $I$ from its distribution $p(I)$ using maximum likelihood~(minimize negative log likelihood): 
\begin{align} 
&p(I) = \begin{cases}
        \prod_{i=1}^L \eta_i^{I(i)}, \quad  \forall i:\quad I(i+1) \in \mathcal{F}(I(i))\\
        0,  \quad \text{else.}
        \end{cases} \\
& I = \operatorname*{argmin}_I \sum_{i=1}^L\text{-log}(\eta_i^{I(i)}), \, \forall i: I(i+1) \in \mathcal{F}(I(i))
\label{eq:decode}
\end{align}
  We build a directed graph $\mathcal{G}$ using $\eta$ and $\mathcal{F}$ as illustrated in Fig.~\ref{fig:tp}. The nodes~(yellow blocks) of $\mathcal{G}$ are connection operations and the input edge cost to a node $c^i_j$ in $\mathcal{G}$ is $\mathrm{-log}(\eta^i_j)$. The path with minimum cost from the source to the sink nodes~(green nodes with gray contour) corresponds to Eq.~\ref{eq:decode}, and we obtained the optimal $I$ using Dijkstra algorithm~\cite{dijkstra1959note}. For cell operations on the selected edges from $I$, we simply use the operation with the largest $\alpha$.
\subsection{Bridging the Discretization Gap}
\label{s:decode}
\begin{figure}
\begin{center}
\includegraphics[width=1\linewidth]{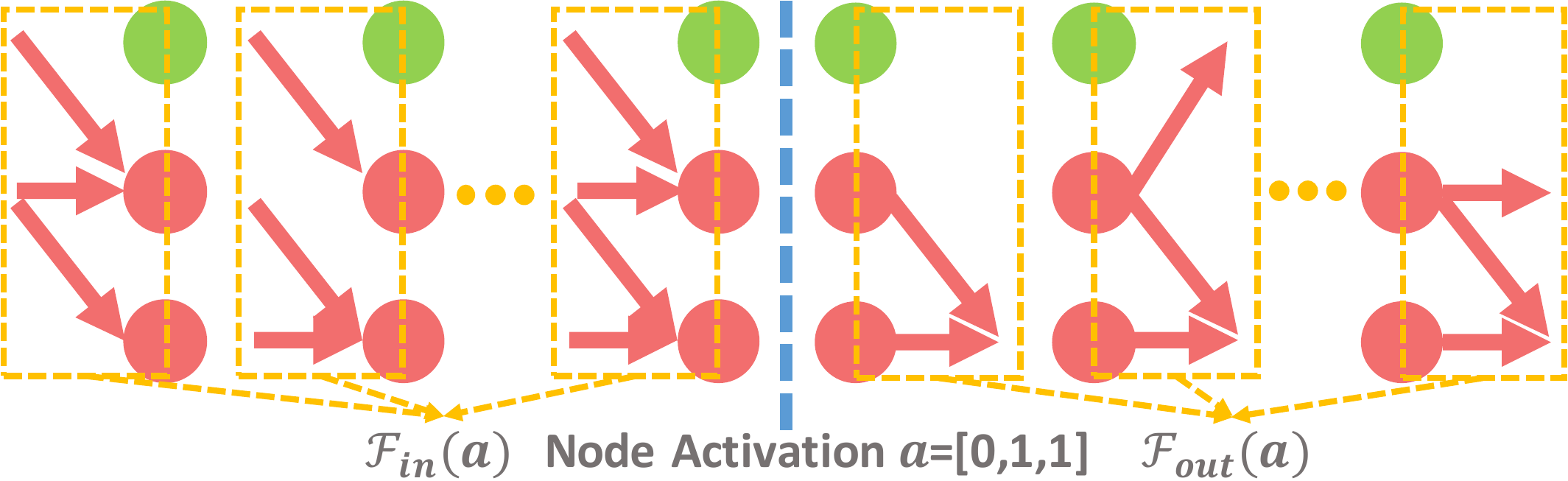}
\end{center}
  \caption{The connection patterns in $\mathcal{F}_{in}(a)$ activates pattern $a$, and all feasible output connection patterns are in $\mathcal{F}_{out}(a)$. $a=[0,1,1]$ means the last two nodes of the super-node are activated.}
\label{fig:tploss}
\end{figure}

To minimize the gap between the continuous representation and the final discretized architecture, we add entropy losses to encourage binarization of $\alpha$ and $\eta$: 
\begin{equation}
\begin{aligned}
\mathcal{L_{\alpha}} &= \frac{-1}{L*E*N}\sum_{i=1}^L\sum_{e=1}^E\sum_{n=1}^N\alpha^{i,e}_n * log(\alpha^{i,e}_n) \\
\mathcal{L_{\eta}} &= \frac{-1}{L*M} \sum_{i=1}^L\sum_{j=1}^M\eta^i_j * log(\eta^i_j) 
\label{eq:entropy}
\end{aligned}
\end{equation}
However, even if the architecture parameters $\alpha$ and $\eta$ are almost binarized, there may still be a large gap due to the topology constraints in the discretization algorithm. Recall the definition of topology feasibility in Sec.~\ref{s:dis}: an activated feature node~(node with at least one input edge) must have an output edge while an in-activated feature node cannot have an output edge. Each super node has $D$ feature nodes, thus there are $2^D-1$ node activation pattern. We define $\mathcal{A}$ as the set of all node activation patterns. Each element $a\in\mathcal{A}$ is a indication function of length $D$, where $a(i)=1$ if the $i$-th node of the super-node is activated. We further define two sets $\mathcal{F}_{in}(a)$ and $\mathcal{F}_{out}(a)$ representing all feasible input and output connection pattern indexes for a super node with node activation $a$ as shown in Fig.~\ref{fig:tploss}. We propose the following topology loss:
\begin{align}
p^i_{in}(a) & = \sum_{j\in {F}_{in}(a)} \eta^i_j, \quad p^i_{out}(a) = \sum_{j\in {F}_{out}(a)} \eta^{i+1}_j \\ \nonumber
\mathcal{L}_{tp} &= -\sum_{i=1}^{L-1}\sum_{a\in \mathcal{A}} (\quad p^i_{in}(a) \text{log}(p^i_{out}(a)) \quad + \\ 
 &\quad\quad\quad (1-p^i_{in}(a)) \text{log}(1-p^i_{out}(a))\quad )
\end{align}
$p^i_{in}(a)$ is the probability that the activation pattern for $s_i$ is $a$, and $p^i_{out}(a)$ is the probability that $s_i$ with pattern $a$ is feasible. By minimizing $\mathcal{L}_{tp}$, the search stage is aware of the topology constraints and encourages all super nodes to be topologically feasible, thus reduce the gap caused by topology constraints in the discretization step.
\subsection{Memory Budget Constraints}
The searched model is usually retrained under different training settings like patch size, filter number, or tasks. Auto-DeepLab~\cite{liu2019auto} used $4\times$ larger image patch and $6\times$ more filters in retraining compared to the search stage. But this can cause out of memory problem for 3D images in retraining, thus we consider memory budget in architecture search. A cell's expected memory usage is estimated by ${M}^{i,e}=\sum_{n=1}^N \alpha^{i,e}_n M_n$. $M_n$ is the memory usage of operation $O_n$~(estimated by tensor size~\cite{gao2020estimating}) defined in Sec.~\ref{s:cell}. The expected memory usage $M_e$ of the searched model is:
\begin{align}
M_e &= \sum_{i=1}^L\sum_{j=1}^M \eta^i_j*(\sum_{e=1}^E M^{i,e} * cp_j(e))
\end{align}
Similar to \cite{li2020learning}, we consider the budget as the percentage $\sigma$ of the maximum memory usage $M_a$, of which all $\alpha$ and $\eta$ equal to one. 
\begin{align}
M_a &= \sum_{i=1}^L\sum_{j=1}^M *(\sum_{e=1}^E (\sum_{n=1}^N M_n) * cp_j(e)) \\
\mathcal{L}_m &= | M_e/M_a - \sigma |_1
\label{eq:mem}
\end{align}
\subsection{Optimization}
We adopt the same optimization strategy as in DARTS~\cite{liu2018darts} and Auto-DeepLab~\cite{liu2019auto}. We partition the training set into \emph{train1} and \emph{train2}, and optimize the network weights $w$~(e.g. convolution kernels) using $\mathcal{L}_{seg}$ on \emph{train1} and network architecture weights $\alpha$ and $p_e$ using $\mathcal{L}_{arch}$ on \emph{train2} alternately. The loss $\mathcal{L}_{seg}$ for $w$ is the evenly sum of dice and cross-entropy loss~\cite{yu2020c2fnas} in segmentation, while
\begin{equation}
    \mathcal{L}_{arch}=\mathcal{L}_{seg}+ t/t_{all}*(\mathcal{L}_{\alpha}+\mathcal{L}_{\eta}+\lambda*\mathcal{L}_{tp}+\mathcal{L}_m)
\end{equation}
$t$ and $t_{all}$ are the current and total iterations for architecture optimization such that the searching is focusing more on $\mathcal{L}_{seg}$ at the starting point. We empirically scale $\mathcal{L}_{tp}$ to the same range with other losses by setting $\lambda$=0.001.
\section{Experiments}
We conduct experiments on the MSD dataset~\cite{simpson2019large} which is a comprehensive benchmark for medical image segmentation. It contains ten segmentation tasks covering different anatomies of interest, modalities and imaging sources~(institutions) and is representative for real clinical problems. Recent C2FNAS~\cite{yu2020c2fnas} reaches state-of-the-art results on MSD dataset using NAS based methods. We follow its experiment settings by searching on the MSD Pancreas dataset and deploying the searched model on all 10 MSD tasks for better comparison. All images are resampled to have a $1.0 \times 1.0 \times 1.0$ $mm^3$ voxel resolution.
\subsection{Implementation Details}
\label{s:imp}
Our search space has $L$=12 layers and $D$=4 resolution levels as shown in Fig.~\ref{fig:ss}. The stem cell at scale 1 has 16 filters and we double the filter number when decreasing the spatial size by half in each axis. The search is conducted on Pancreas dataset following the same 5 fold data split~(4 for training and last 1 for validation) as C2FNAS~\cite{yu2020c2fnas}. We use SGD optimizer with momentum 0.9, weight decay of 4e-5 for network weights $w$. We train $w$ for the first one thousand~(1k) warm-up and following 10k iterations without updating architecture. The architecture weights $\alpha, p_e$ are initialized with Gaussian $\mathcal{N}(1,0.01), \mathcal{N}(0,0.01)$ respectively before softmax and sigmoid. In the following 10k iterations, we jointly optimize $w$ with SGD and $\alpha, p_e$ with Adam optimizer~\cite{kingma2014adam}~(learning rate 0.008, weight decay 0). The learning rate of SGD linearly increases from 0.025 to 0.2 in the first 1k warm-up iterations, and decays with factor 0.5 at the following [8k, 16k] iterations. The search is conducted on 8 GPUs with batch size 8~(each GPU with one 96$\times$96$\times$96 patch). The patches are randomly augmented with 2D rotation by [90, 180, 270] degrees in the x-y plane and flip in all three axis. The total training iterations, SGD learning rate scheduler and data pre-processing and augmentation are the same with C2FNAS~\cite{yu2020c2fnas}.  After searching, the discretized model is randomly initialized and retrained with doubled filter number and doubled batch size to match C2FNAS~\cite{yu2020c2fnas}'s setting. We use the SGD optimizer with 1k warm-up and 40k training iterations and decay the learning rate by a factor of 0.5 at [8k, 16k, 24k, 32k] iterations after warm-up. The learning rate scheduler is the same with search stage in the warm-up and the first 20k iterations. The latter 20k iterations are for better convergence and match the 40k total retraining iterations used in C2FNAS~\cite{yu2020c2fnas}. The same data augmentation as C2FNAS~(also the same as the search stage) is used for the Pancreas dataset for better comparison. To test the generalizability of the searched model, we retrain the model on all of the rest nine tasks. Some tasks in the MSD dataset contain very few training data so we use additional basic 2D data augmentations of random rotation, scaling and gamma correction for all nine tasks. We use patch size $96 \times 96 \times 96$ and stride $16 \times 16 \times 16$ for all ten tasks except Prostate and Hippocampus. Prostate data has very few slices~(less than 40) in the z-axis, so we use patch size $96 \times 96 \times 32$ and stride $16 \times 16 \times 4$. Hippocampus data size is too small~(around $36 \times 50 \times 35$) and we use patch size $32 \times 32  \times 32$ and stride $4 \times 4 \times 4$. Post-processing with largest connected component is also applied.

\begin{table}[]
\centering
\caption{Comparison of FLOPs, Parameters and Retraining GPU memory usage and the
5-Fold cross validation Dice-Sørensen score of our searched architectures on Pancreas dataset}
\label{t:search}
\resizebox{\columnwidth}{!}{%
\begin{tabular}{l|c|c|c|ccc}
\hline
Model          & \begin{tabular}[c]{@{}c@{}}FLOPs\\  (G)\end{tabular} & \begin{tabular}[c]{@{}c@{}}Params. \\ (M)\end{tabular} & \begin{tabular}[c]{@{}c@{}}Memory\\  (MB)\end{tabular} & DSC1  & DSC2  & Avg.  \\ \hline
3D UNet~\cite{cciccek20163d}~(nn-UNet)        & 658                                                  & 18                                                    & 9176                                                   & -     & -     & -     \\ \hline
Attention UNet~\cite{oktay2018attention} & 1163                                                 & 104                                                   & 13465                                                  & -     & -     & -     \\ \hline
C2FNAS~\cite{yu2020c2fnas}         & 151                                                  & 17                                                    & 5730                                                   & -     & -     & -     \\ \hline
DiNTS~($\sigma$=0.2)      & 146                                                  & 163                                                   & 5787                                                   & 77.94 & 48.07 & 63.00 \\ \hline
DiNTS~($\sigma$=0.5)      & 308                                                  & 147                                                   & 10110                                                  & \textbf{80.20} & 52.25 & 66.23 \\ \hline
DiNTS~($\sigma$=0.8)      & 334                                                  & 152                                                   & 13018                                                  & 80.06 & \textbf{52.53} & \textbf{66.29} \\ \hline
\end{tabular}
}
\end{table}
\begin{figure}[t]
\begin{center}
\includegraphics[width=1\linewidth]{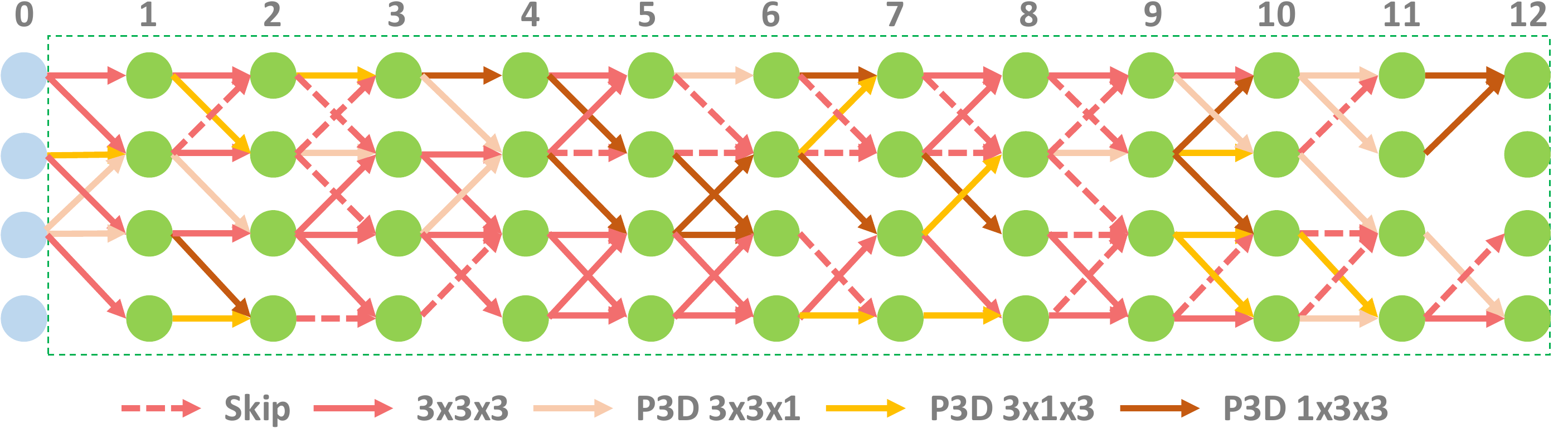}
(a) Searched architecture with $\sigma=0.8$
\includegraphics[width=1\linewidth]{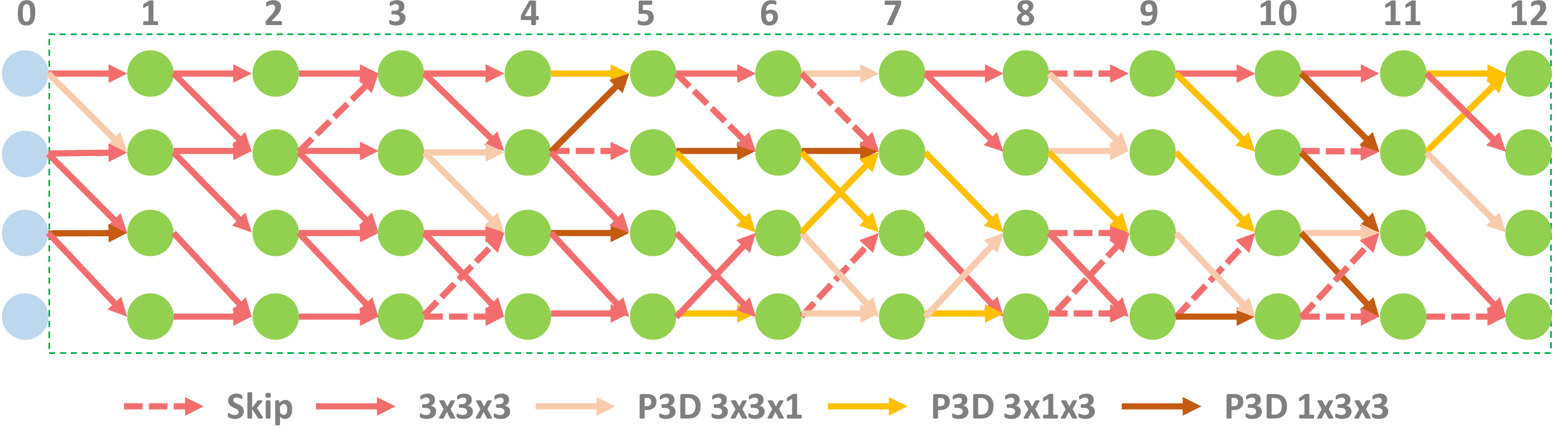}
(b) Searched architecture with $\sigma=0.5$
\includegraphics[width=1\linewidth]{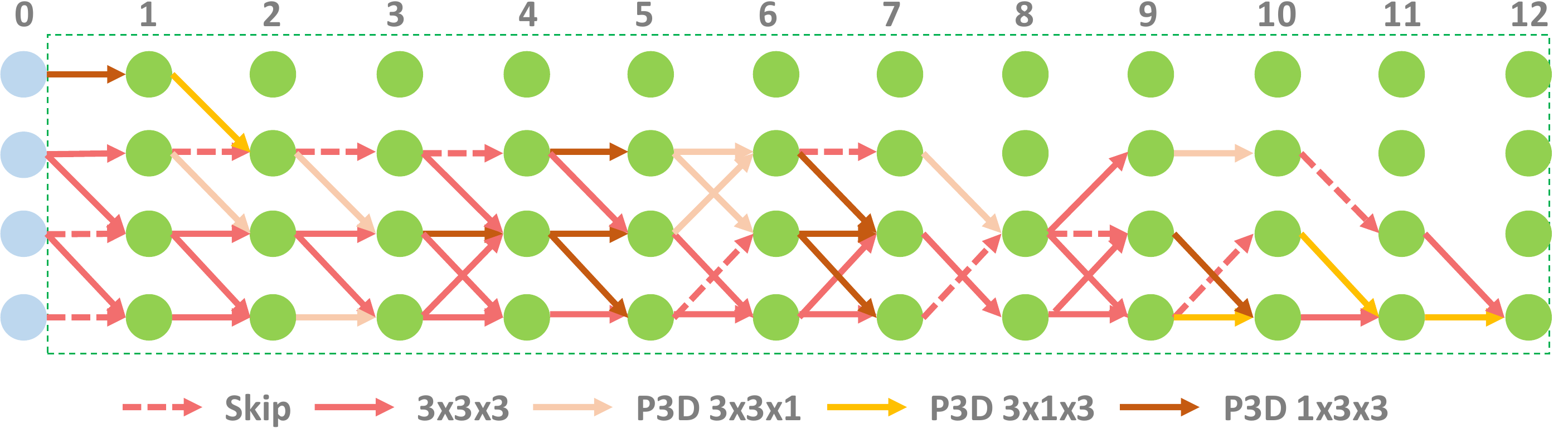}
(c) Searched architecture with $\sigma=0.2$
\end{center}
  \caption{Searched architectures~(not including the stem in Fig.~\ref{fig:ss}) on Pancreas dataset with varying memory constraints.}
\label{fig:s_result}
\end{figure}
\subsection{Pancreas Dataset Search Results}
\label{sec:search-result}
The search takes 5.8 GPU days while C2FNAS takes 333 GPU days on the same dataset~(both using 8 16GB V100 GPU). We vary the memory constraints $\sigma=[0.2,0.5,0.8]$ and show the search results in Fig.~\ref{fig:s_result}. The searched models have highly flexible topology which are searched jointly with the cell level. The 5-fold cross-validation results on Pancreas are shown in Table~\ref{t:search}. By increasing $\sigma$, the searched model is more ``dense in connection'' and can achieve better performance while requiring more GPU memory~(estimated using PyTorch~\cite{paszke2019pytorch} functions in training described in Sec.~\ref{s:imp}). The marginal performance drop by decreasing $\sigma=0.8$ to $\sigma=0.5$ shows that we can reduce memory usage without losing too much accuracy. Although techniques like mixed-precision training~\cite{micikevicius2017mixed} can be used to further reduce memory usage, our memory aware search tries to solve this problem from NAS perspective. Compared to nnUNet~\cite{isensee2019nnunet}~(represented by 3D UNet because it ensembles 2D/3D/cascaded-3D U-Net differently for each task) and C2FNAS in Table~\ref{t:search}, our searched models have no advantage in FLOPs and Parameters which are important in mobile settings. We argue that for medical image analysis, light model and low latency are less a focus than better GPU memory usage and accuracy. Our DiNTS can optimize the usage of the available GPU and achieve better performance.

\subsection{Segmentation Results on MSD} 
\label{sec:test-result}
The searched model with $\sigma=0.8$ from Pancreas is used for retraining and testing on all ten tasks of MSD dataset. Similar to the model ensemble used in nnUNet~\cite{isensee2019nnunet} and C2FNAS~\cite{yu2020c2fnas}, we use a 5 fold cross validation for each task and ensemble the results using majority voting. The largest connected component post-processing in nnUNet~\cite{isensee2019nnunet} is also applied. The Dice-Sørensen~(DSC) and Normalised Surface Distance~(NSD) as used in the MSD challenge are reported for the test set in Table~\ref{t:result}. nnUNet~\cite{isensee2019nnunet} uses extensive data augmentation, different hyper-parameters like patch size, batch size for each task and ensembles networks with different architectures. It focuses on hyper-parameter selection based on hand-crafted rules and is the champion of multiple medical segmentation challenges including MSD. Our method and C2FNAS~\cite{yu2020c2fnas} focus on architecture search and use consistent hyper-parameters and basic augmentations for all ten tasks. We achieved better results than C2FNAS~\cite{yu2020c2fnas} in all tasks with similar hyper-parameters while only takes 1.7\% searching time. Comparing to nn-UNet~\cite{isensee2019nnunet}, we achieve much better performance on challenging datasets like Pancrease, Brain and Colon, while worse on smaller datasets like Heart~(10 test cases), Prostate~(16 test cases) and Spleen~(20 test cases). Task-specific hyper-parameters, test-time augmentation, extensive data augmentation and ensemble more models as used in nn-UNet~\cite{isensee2019nnunet} might be more effective on those small datasets than our unified DiNTS searched architecture. Overall, we achieved the best average results and top ranking in the MSD challenge leaderboard, showing that a non-UNet based topology can achieve superior performance in medical imaging. 
\begin{table}[]
\resizebox{\columnwidth}{!}{%
\begin{tabular}{l|cccc|cccc}
\hline
       & \multicolumn{8}{c}{Brain}                                                           \\ \hline
Metric & DSC1  & DSC2  & DSC3  & Avg.           & NSD1  & NSD2  & NSD3  & Avg.   \\
\hline
CerebriuDIKU~\cite{perslev2019one} & \textbf{69.52}&	43.11&	66.74&	59.79&		88.25	&68.98&	88.90&	82.04\\
NVDLMED~\cite{xia20203d} & 67.52   & 45.00  & 68.01 & 60.18 & 86.99 & 69.77         & 89.82         & 82.19         \\
Kim et al~\cite{kim2019scalable} & 67.40 & 45.75 &68.26 & 60.47	& 86.65 & 72.03 & 90.28	& 82.99           \\
nnUNet~\cite{isensee2019nnunet} & 68.04 & 46.81 & 68.46 & 61.10          & 87.51 & 72.47 & 90.78 & 83.59 \\
C2FNAS~\cite{yu2020c2fnas} & 67.62 & 48.60 & 69.72 & 61.98          & 87.61 & 72.87 & 91.16 & 83.88 \\ \hline
DiNTS   & 69.28 & \textbf{48.65} & \textbf{69.75} & \textbf{62.56} & \textbf{89.33} & \textbf{73.16} & \textbf{91.69} & \textbf{84.73} \\
\hline 

\end{tabular}
}

\vspace{0.5\baselineskip}
\resizebox{\columnwidth}{!}{%
\begin{tabular}{l|cc|ccc|ccc}
\hline
       & \multicolumn{2}{c|}{Heart} & \multicolumn{6}{c}{Liver}                                                                   \\ \hline
Metric & DSC1            & NSD1            & DSC1           & DSC2           & Avg.           & NSD1           & NSD2           & Avg.           \\ \hline
CerebriuDIKU~\cite{perslev2019one} & 89.47	&	90.63 	&94.27	&	57.25		&	75.76	&		96.68	&	72.60	&		84.64 \\
NVDLMED~\cite{xia20203d} & 92.46   & 95.57  &95.06 & 71.40 & 83.23 & 98.26          & 87.16          & 92.71           \\
Kim et al~\cite{kim2019scalable} & 93.11   & 96.44 & 94.25&	72.96&		83.605	&	96.76&	88.58	&	92.67          \\
nnUNet~\cite{isensee2019nnunet} & \textbf{93.30}   & \textbf{96.74}  & \textbf{95.75} & \textbf{75.97} & \textbf{85.86} & 98.55          & 90.65          & 94.60           \\
C2FNAS~\cite{yu2020c2fnas} & 92.49           & 95.81           & 94.98          & 72.89          & 83.94          & 98.38          & 89.15          & 93.77          \\ \hline
DiNTS   & 92.99           & 96.35           & 95.35          & 74.62          & 84.99          & \textbf{98.69} & \textbf{91.02} & \textbf{94.86} \\ \hline
\end{tabular}
}

\vspace{0.5\baselineskip}
\resizebox{\columnwidth}{!}{%
\begin{tabular}{l|cc|ccc|ccc}
\hline
       & \multicolumn{2}{c|}{Lung}       & \multicolumn{6}{c}{Hippocampus}                                                                    \\ \hline
Metric & DSC1           & NSD1           & DSC1           & DSC2           & Avg.           & NSD1           & NSD2           & Avg.           \\ \hline
CerebriuDIKU~\cite{perslev2019one} & 58.71 & 56.10 & 89.68 &	88.31 &		89.00	&	97.42 &	97.42	&	97.42 \\
NVDLMED~\cite{xia20203d}  & 52.15          & 50.23          & 87.97 & 86.71 & 87.34 & 96.07 & 96.59          & 96.33 \\
Kim et al~\cite{kim2019scalable}& 63.10 &62.51 &90.11&	88.72	&	89.42	&	97.77&	\textbf{97.73}	&	\textbf{97.75} \\
nnUNet~\cite{isensee2019nnunet} & 73.97          & 76.02          & \textbf{90.23} & \textbf{88.69} & \textbf{89.46} & \textbf{97.79} & 97.53          & 97.66 \\
C2FNAS~\cite{yu2020c2fnas} & 70.44          & 72.22          & 89.37          & 87.96          & 88.67         & 97.27          & 97.35          & 97.31          \\ \hline
DiNTS   & \textbf{74.75} & \textbf{77.02} & 89.91          & 88.41          & 89.16          & 97.76          & 97.56 & 97.66  \\ \hline
\end{tabular}
}


\vspace{0.5\baselineskip}
\resizebox{\columnwidth}{!}{%
\begin{tabular}{l|cc|ccc|ccc}
\hline
       & \multicolumn{2}{c|}{Spleen}     & \multicolumn{6}{c}{Prostate}                                                                       \\ \hline
Metric & DSC1           & NSD1           & DSC1           & DSC2           & Avg.           & NSD1           & NSD2           & Avg.           \\ \hline
CerebriuDIKU~\cite{perslev2019one} & 95.00 &98.00 &69.11&	86.34	&	77.73		&94.72	&97.90  &	96.31 \\
NVDLMED~\cite{xia20203d} & 96.01          & 99.72          & 69.36          & 86.66          & 78.01          & 92.96          & 97.45          & 95.21          \\ 
Kim et al~\cite{kim2019scalable} & 91.92 & 94.83 & 72.64 & 	89.02	&	80.83	&	95.05 &	98.03	&	96.54 \\
nnUNet~\cite{isensee2019nnunet} & \textbf{97.43} & \textbf{99.89} & \textbf{76.59} & \textbf{89.62} & \textbf{83.11} & \textbf{96.27} & \textbf{98.85} & \textbf{97.56} \\
C2FNAS~\cite{yu2020c2fnas} & 96.28          & 97.66          & 74.88          & 88.75          & 81.82          & 98.79          & 95.12          & 96.96          \\ \hline
DiNTS   & 96.98          & 99.83          & 75.37          & 89.25          & 82.31          & 95.96          & 98.82          & 97.39          \\ \hline
\end{tabular}
}


\vspace{0.5\baselineskip}
\resizebox{\columnwidth}{!}{%
\begin{tabular}{l|cc|ccc|ccc}
\hline
       & \multicolumn{2}{c|}{Colon}      & \multicolumn{6}{c}{Hepatic Vessels}                                                                \\ \hline
Metric & DSC1           & NSD1           & DSC1           & DSC2           & Avg.           & NSD1           & NSD2           & Avg.           \\ \hline
CerebriuDIKU~\cite{perslev2019one} & 28.00 & 43.00 & 59.00 &	38.00&		48.50&		79.00 &	44.00&		61.50 \\
NVDLMED~\cite{xia20203d} & 55.63 & 66.47 & 61.74 & 61.37 & 61.56 & 81.61 & 68.82 & 75.22 \\
Kim et al~\cite{kim2019scalable} & 49.32& 62.21 &62.34	&68.63&		65.485	&	83.22&	78.43	&	80.825 \\
nnUNet~\cite{isensee2019nnunet} & 58.33          & 68.43          & \textbf{66.46} & \textbf{71.78} & \textbf{69.12} & \textbf{84.43} & 80.72          & \textbf{82.58} \\
C2FNAS~\cite{yu2020c2fnas} & 58.90           & \textbf{72.56} & 64.30          & 71.00          & 67.65          & 83.78          & 80.66          & 82.22          \\ \hline
DiNTS  & \textbf{59.21} & 70.34          & 64.50          & 71.76          & 68.13          & 83.98          & \textbf{81.03} & 82.51          \\ \hline
\end{tabular}
}


\vspace{0.5\baselineskip}
\resizebox{\columnwidth}{!}{%
\begin{tabular}{l|ccc|ccc|cc}
\hline
       & \multicolumn{6}{c|}{Pancreas}                                                                       & \multicolumn{2}{c}{\textbf{Overall}} \\ \hline
Metric & DSC1           & DSC2           & Avg.           & NSD1           & NSD2           & Avg.           & DSC               & NSD               \\ \hline
CerebriuDIKU~\cite{perslev2019one} & 71.23 &	24.98	&	48.11	&	91.57	&46.43	&	69.00 & 67.01& 77.86 \\
NVDLMED~\cite{xia20203d}  & 77.97 & 44.49         & 61.23         & 94.43          & 63.45         & 78.94         & 72.78             & 83.26            \\
Kim et al~\cite{kim2019scalable} & 80.61 &	51.75	&	66.18	&	95.83&	73.09	&	84.46 & 74.34 & 85.12 \\
nnUNet~\cite{isensee2019nnunet} & \textbf{81.64} & 52.78          & 67.21          & 96.14          & 71.47          & 83.81          & 77.89             & 88.09             \\
C2FNAS~\cite{yu2020c2fnas} & 80.76          & 54.41          & 67.59          & 96.16          & 75.58          & 85.87          & 76.97             & 87.83             \\ \hline
DiNTS  & 81.02          & \textbf{55.35} & \textbf{68.19} & \textbf{96.26} & \textbf{75.90} & \textbf{86.08} & \textbf{77.93}    & \textbf{88.68}    \\ \hline
\end{tabular}
}
\caption{Dice-Sørensen score~(DSC) and Normalised Surface Distance~(NSD) results on the MSD test dataset~(numbers from MSD challenge live leaderboard).}
\label{t:result}
\end{table}
\subsection{Ablation Study}
\subsubsection{Search on Different Datasets}
The models in Sec.~\ref{sec:search-result} and Sec.~\ref{sec:test-result} are searched from the Pancreas dataset~(282 CT 3D training images). To test the generalizability of DiNTS, we perform the same search as in Sec.~\ref{s:imp} on Brain~(484 MRI data), Liver~(131 CT data) and Lung~(64 CT data) covering big, medium and small datasets. The results are shown in Table.~\ref{t:other} and demonstrate the good generalizability of our DiNTS. 
\begin{table}[]
\resizebox{\columnwidth}{!}{%
\begin{tabular}{c|c|c|c|c|cc}
\hline
Test Dataset   & \multicolumn{2}{c|}{Brain}                       & \multicolumn{2}{c|}{Liver} & \multicolumn{2}{c}{Lung}                       \\ \hline
Search Dataset & Brain                                 & Pancreas & Liver           & Pancreas & \multicolumn{1}{c|}{Lung}           & Pancreas \\ \hline
DSC1           & \textbf{80.20}                        & 79.68    & \textbf{94.15}  & 94.12    & \multicolumn{1}{c|}{\textbf{69.30}} & 68.90    \\ \hline
DSC2           & \textbf{61.09} & 60.67    & \textbf{58.74}  & 57.86    & \multicolumn{1}{c|}{-}     & -        \\ \hline
DSC3           & \textbf{77.63}                        & 77.48    & -     & -        & \multicolumn{1}{c|}{-}     & -        \\ \hline
Avg.           & \textbf{72.97}                        & 72.61    & \textbf{76.44}  & 75.99    & \multicolumn{1}{c|}{\textbf{69.30}} & 68.90    \\ \hline
\end{tabular}
}
\caption{Dice-Sørensen score~(DSC) of 5-fold cross validation on Brain, Liver and Lung datasets of architectures searched from Pancreas, Brain, Liver and Lung datasets with $\sigma=0.8$.}
\label{t:other}
\end{table}

\subsubsection{Necessity of Topology Loss}
\begin{figure}
\begin{center}
\includegraphics[width=0.8\linewidth]{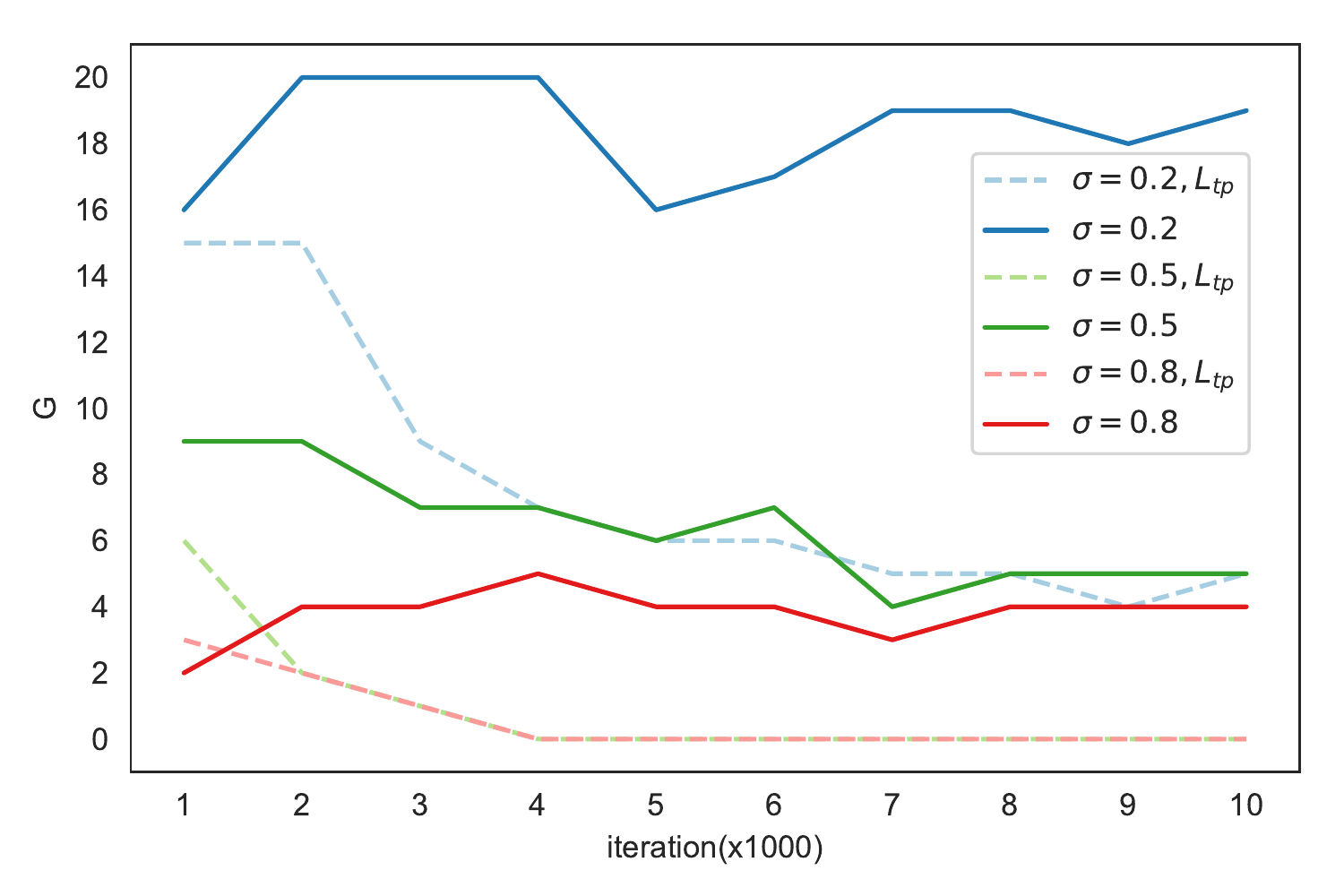}
\end{center}
  \caption{The indication $G$ of discretization gap during architecture search with different memory constraints $\sigma$. With topology loss~(dashed line),  $G$ is decreased compared to no topology loss~(solid line), showing the importance of topology loss.}
\label{fig:g}
\end{figure}

As illustrated in Sec.~\ref{s:intro}, the discretization algorithm discards topologically infeasible edges~(even with large probabilities), which causes a gap between feature flow in the optimized continuous model~(Eq.~\ref{eq:flow})
and the discrete model. Our topology loss encourages connections with large probabilities to be feasible, thus will not be discarded and causing the gap. 
We denote $C_{max}$ as the topology decoded by selecting connection $j$ with largest $\eta^i_j$ for each layer $i$~(can be infeasible). $C_{top}$ is the topology decoded by our discretization algorithm. $C_{max}, C_{top}$ are the indication matrices of size [$L,E$] representing whether an edge is selected, and $G=\sum_{i=1}^L\sum_{e=1}^E |C_{max}(i,e)-C_{top}(i,e)|$. Larger $G$ represents larger gap between the feature flow before and after discretization. Fig.~\ref{fig:g} shows the change of $G$ during search with/without topology loss under different memory constraints. With topology loss, the gap between $C_{max}$ and $C_{top}$ is reduced, and it's more crucial for smaller $\sigma$ where the searched architecture is more sparse and more likely to have topology infeasibility. 

\section{Conclusions}
In this paper, we present a novel differentiable network topology search framework~(DiNTS) for 3D medical image segmentation. By converting the feature nodes with varying spatial resolution into super nodes, we are able to focus on connection patterns rather than individual edges, which enables more flexible network topologies and a discretization aware search framework. Medical image segmentation challenges have been dominated by U-Net based architectures~\cite{isensee2019nnunet}, even NAS-based C2FNAS is searched within a U-shaped space. DiNTS's topology search space is highly flexible and achieves the best performance on the benchmark MSD challenge using non-UNet architectures, while only taking 1.7\% search time compared to C2FNAS. Since directly converting Auto-DeepLab~\cite{liu2019auto} to the 3D version will have memory issues, we cannot fairly compare with it. For future work, we will test our proposed algorithm on 2D natural image segmentation benchmarks and explore more complex cells.
{\small
\bibliographystyle{ieee_fullname}
\bibliography{egbib}
}


\end{document}